\documentclass[runningheads]{llncs}

\usepackage{eccv}

\usepackage{eccvabbrv}

\usepackage{graphicx}
\usepackage{booktabs}
\usepackage{multirow}
\usepackage{multicol}
\usepackage{amsmath}
\usepackage{amssymb}
\usepackage{xcolor}
\usepackage{colortbl}
\usepackage{marvosym}

\usepackage[accsupp]{axessibility}  

\usepackage{hyperref}

\usepackage{orcidlink}

\begin{document}

\title{XAttnRes: Cross-Stage Attention Residuals for Medical Image Segmentation}

\titlerunning{XAttnRes for Medical Image Segmentation}

\author{Xinyu Liu$^{\dag}$\inst{1} \and
Qing Xu$^{\dag}$\inst{2} \and
Zhen Chen\textsuperscript{\Letter}\inst{3}}

\authorrunning{Liu, Xu, and Chen}

\institute{Imperial College London \and University of Nottingham \and
The Hong Kong Polytechnic University\\
\email{z.chen@polyu.edu.hk}}

\maketitle

{
  \renewcommand{\thefootnote}{}  \footnotetext{\hspace{-4pt}$^{\dag}$Equal Contribution}
}

\begin{abstract}
In the field of Large Language Models (LLMs), Attention Residuals have recently demonstrated that learned, selective aggregation over all preceding layer outputs can outperform fixed residual connections. We propose \textbf{Cross-Stage Attention Residuals (XAttnRes)}, a mechanism that maintains a global feature history pool accumulating both encoder and decoder stage outputs. Through lightweight pseudo-query attention, each stage selectively aggregates from all preceding representations. To bridge the gap between the same-dimensional Transformer layers in LLMs and the multi-scale encoder-decoder stages in segmentation networks, XAttnRes introduces spatial alignment and channel projection steps that handle cross-resolution features with negligible overhead. When added to existing segmentation networks, XAttnRes consistently improves performance across four datasets and three imaging modalities. We further observe that XAttnRes alone, even without skip connections, achieves performance on par with the baseline, suggesting that learned aggregation can recover the inter-stage information flow traditionally provided by predetermined connections.
\end{abstract}


\section{Introduction}

Segmentation, the fundamental task of partitioning an image into semantically meaningful regions, is a cornerstone of visual analysis~\cite{long2015fcn, badrinarayanan2017segnet, xie2021segformer}. It is particularly critical in medical imaging, where accurate pixel-level delineation of organs, lesions, and tissues directly impacts clinical decision-making~\cite{litjens2017survey}. In this field, encoder-decoder architectures have become the dominant paradigm~\cite{ronneberger2015unet, isensee2021nnunet}. The encoder extracts hierarchical feature representations through progressive downsampling, the decoder recovers spatial resolution, and skip connections forward encoder features to decoder stages to preserve fine-grained details. Since U-Net~\cite{ronneberger2015unet} popularized this design, a rich line of work has improved the encoder~\cite{chen2021transunet, cao2022swinunet, hatamizadeh2022unetr}, the decoder~\cite{rahman2024emcad, rahman2023cascade}, and the skip connection itself~\cite{oktay2018attention, zhou2018unetpp, huang2020unet3p, wang2022uctransnet}.

However, across all these advances, feature routing between stages remains architecturally predetermined: the topology of which features reach which stages is designed by the architect, not learned from data. A parallel development in large language models suggests this need not be the case. \textbf{Attention Residuals (AttnRes)}~\cite{chen2026attnres} recently showed that the fixed residual connections in Transformers, which accumulate layer outputs with uniform unit weights, can be replaced by learned softmax attention over all preceding layer outputs. A single pseudo-query vector per layer selects which earlier representations to aggregate, enabling content-aware, position-specific routing with minimal overhead. This raises a natural question: \emph{can the same principle of learned aggregation benefit the multiple stages in medical image segmentation?}

Adapting AttnRes to segmentation networks is non-trivial. In large language models, transformer layers share the same dimensionality, so the original mechanism operates on uniform-shaped representations. Encoder-decoder stages, by contrast, produce features at different spatial resolutions and channel widths, and information must flow not only from encoder to decoder but also across decoder stages at multiple scales. We address these challenges with a simple mechanism named \textbf{Cross-Stage Attention Residuals (XAttnRes)}, which extends AttnRes to handle multi-scale, cross-resolution feature routing. XAttnRes maintains a global feature history pool that accumulates both encoder and decoder stage outputs, aligns them to a common resolution through lightweight spatial pooling and channel projection, and applies pseudo-query attention for selective aggregation. This enables each stage to draw upon all preceding representations in the network, providing a learned routing path that operates alongside existing architectural components. We validate XAttnRes on two representative architectures, the classical U-Net \cite{ronneberger2015unet} and the state-of-the-art EMCAD~\cite{rahman2024emcad}, across four datasets spanning three imaging modalities.

\begin{figure}[!tb]
    \centering
    \includegraphics[width=\textwidth]{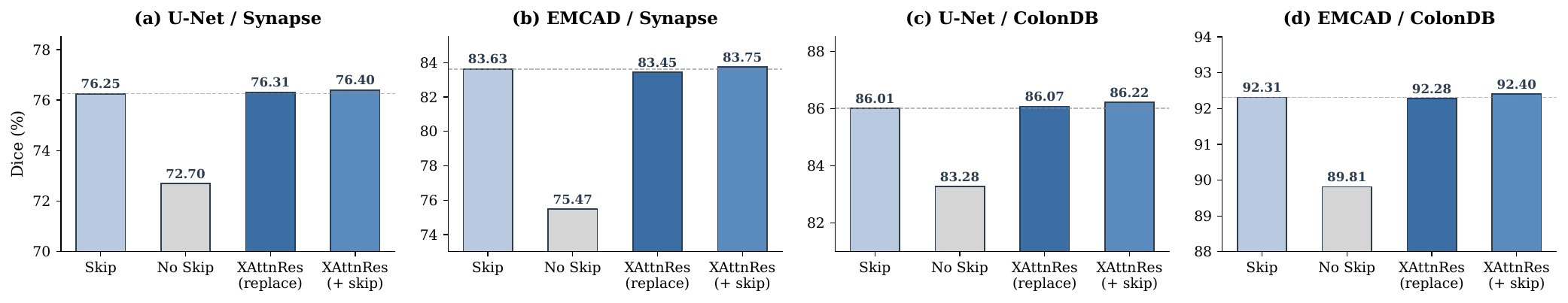}
    \caption{Effect of XAttnRes across two backbones (U-Net and EMCAD) and two benchmarks (Synapse multi-organ CT and ColonDB polyp segmentation). Adding XAttnRes alongside existing architectures (``XAttnRes + skip'') consistently improves over the baseline. Removing skip connections (``No Skip'') degrades performance, but XAttnRes alone (``replace'') recovers most of this drop. Dashed lines indicate the baseline.}
    \label{fig:bar_chart}
\end{figure}

As shown in Figure~\ref{fig:bar_chart}, XAttnRes consistently improves segmentation performance when added to existing architectures. Our contributions are:
\begin{itemize}
    \item We are, to the best of our knowledge, \textbf{the first to adapt Attention Residuals from large language models to medical image segmentation}, extending the mechanism to handle cross-resolution, multi-scale feature routing in encoder-decoder architectures.
    \item We propose \textbf{XAttnRes}, which maintains a global feature history pool across encoder and decoder stages and enables content-aware, per-position feature aggregation with negligible parameter overhead.
    \item We validate XAttnRes on four datasets across three imaging modalities, showing consistent improvements when added to both vanilla U-Net and the state-of-the-art EMCAD framework. We further observe that \textbf{XAttnRes alone can match the performance of predetermined skip connections}, highlighting its potential as a learned alternative.
\end{itemize}

\section{Related Work}

\subsection{Feature routing in encoder-decoder segmentation}

Encoder-decoder architectures~\cite{long2015fcn, ronneberger2015unet} form the backbone of medical image segmentation. Information routing between stages has been a central design consideration: U-Net~\cite{ronneberger2015unet} introduced skip connections, Attention U-Net~\cite{oktay2018attention} added spatial attention gates, UNet++~\cite{zhou2018unetpp} inserted nested dense pathways, UNet 3+~\cite{huang2020unet3p} introduced full-scale cross-resolution connections, and UCTransNet~\cite{wang2022uctransnet} applied channel Transformers for cross-scale alignment. Encoder designs have evolved from CNNs~\cite{he2016resnet} to Transformers~\cite{chen2021transunet, cao2022swinunet, hatamizadeh2022unetr, huang2022missformer}, and decoder designs have incorporated efficient multi-scale attention~\cite{rahman2024emcad, rahman2023cascade}. In all cases, the routing topology between stages remains architecturally predetermined~\cite{wang2024udtransnet, ibtehaz2020multiresunet, peng2025unetv2, luo2025rethinkingunet, zhang2024fafs}.

In general computer vision, cross-layer feature aggregation has been explored through DenseNet~\cite{huang2017densenet}, Feature Pyramid Networks~\cite{lin2017fpn}, HRNet~\cite{sun2019hrnet}, Highway Networks~\cite{srivastava2015highway}, and Non-local Networks~\cite{wang2018nonlocal}. These provide various forms of multi-scale or cross-depth feature reuse, but their routing topologies are likewise fixed at design time. Our work introduces a fully learned, per-position routing mechanism inspired by recent advances in LLMs, which can both enhance existing routing and, as we show, serve as a standalone alternative for feature aggregation.

\subsection{Attention Residuals in large language models}

Attention Residuals (AttnRes) were proposed by Chen et al.~\cite{chen2026attnres} for Transformer-based LLMs. The standard residual connection accumulates layer outputs with fixed unit weights: $\mathbf{h}_l = \mathbf{h}_{l-1} + f_l(\mathbf{h}_{l-1})$. As network depth grows, this causes two problems: feature dilution, where each layer's relative contribution to the accumulated sum diminishes, and unbounded magnitude growth, a well-known issue in PreNorm Transformers. AttnRes replaces the fixed accumulation with softmax attention over all preceding layer outputs, parameterized by a single learned pseudo-query $\mathbf{w}_l \in \mathbb{R}^d$ per layer. This enables selective, content-aware access to earlier representations while keeping output magnitudes bounded.

While effective, Full AttnRes requires storing all layer outputs, which incurs large memory consumption. The practical Block AttnRes variant addresses this by partitioning layers into blocks: within each block, standard residuals are used, and between blocks, attention-based aggregation is applied. This reduces memory while recovering most of Full AttnRes's gains. A key design choice shared by both variants is zero-initialization of all pseudo-queries, which causes the mechanism to start as uniform averaging and gradually specialize during training.

We extend Attention Residuals from \emph{same-dimensional layer aggregation} in LLMs to \emph{cross-resolution, cross-stage aggregation} in segmentation networks. This extension introduces an additional challenge: features in a U-Net have different spatial resolutions and channel dimensions at each stage, requiring a lightweight alignment step before attention can be applied. We address this through adaptive spatial pooling and $1\times1$ convolution projections, preserving the simplicity and low overhead of the original AttnRes mechanism.

\section{Method}

We first review Attention Residuals in the LLM context (Section~\ref{sec:prelim}), then present our cross-stage extension for segmentation (Section~\ref{sec:crossattnres}), and finally describe how XAttnRes integrates into existing architectures as a versatile plug-in (Section~\ref{sec:plugin}).

\subsection{Preliminaries: Attention Residuals}
\label{sec:prelim}

In standard Transformers~\cite{vaswani2017attention}, the residual connection at layer $l$ accumulates outputs with a fixed unit weight:
\begin{equation}
    \mathbf{h}_l = \mathbf{h}_{l-1} + f_l(\mathbf{h}_{l-1}),
    \label{eq:standard_res}
\end{equation}
where $f_l$ denotes the layer computation (e.g., self-attention or feed-forward network). Attention Residuals~\cite{chen2026attnres} replace this fixed accumulation with a learned, selective aggregation over all preceding layer outputs:
\begin{equation}
    \mathbf{h}_l = \sum_{i=0}^{l-1} \alpha_{i \to l} \cdot \mathbf{V}_i, \quad \alpha_{i \to l} = \frac{\exp\!\big(\mathbf{w}_l^\top \cdot \text{RMSNorm}(\mathbf{V}_i)\big)}{\sum_{j=0}^{l-1} \exp\!\big(\mathbf{w}_l^\top \cdot \text{RMSNorm}(\mathbf{V}_j)\big)},
    \label{eq:attnres}
\end{equation}
where $\mathbf{w}_l \in \mathbb{R}^d$ is a learned pseudo-query initialized to zero, and $\mathbf{V}_i$ is the output of layer $i$. All representations share the same shape $[B, T, D]$, and the attention operates independently per token position. The softmax normalization bounds output magnitudes regardless of depth, and zero-initialization ensures the mechanism starts as uniform averaging before specializing during training.

\subsection{Cross-Stage Attention Residual}
\label{sec:crossattnres}

\begin{figure}[!tb]
    \centering
    \includegraphics[width=\textwidth]{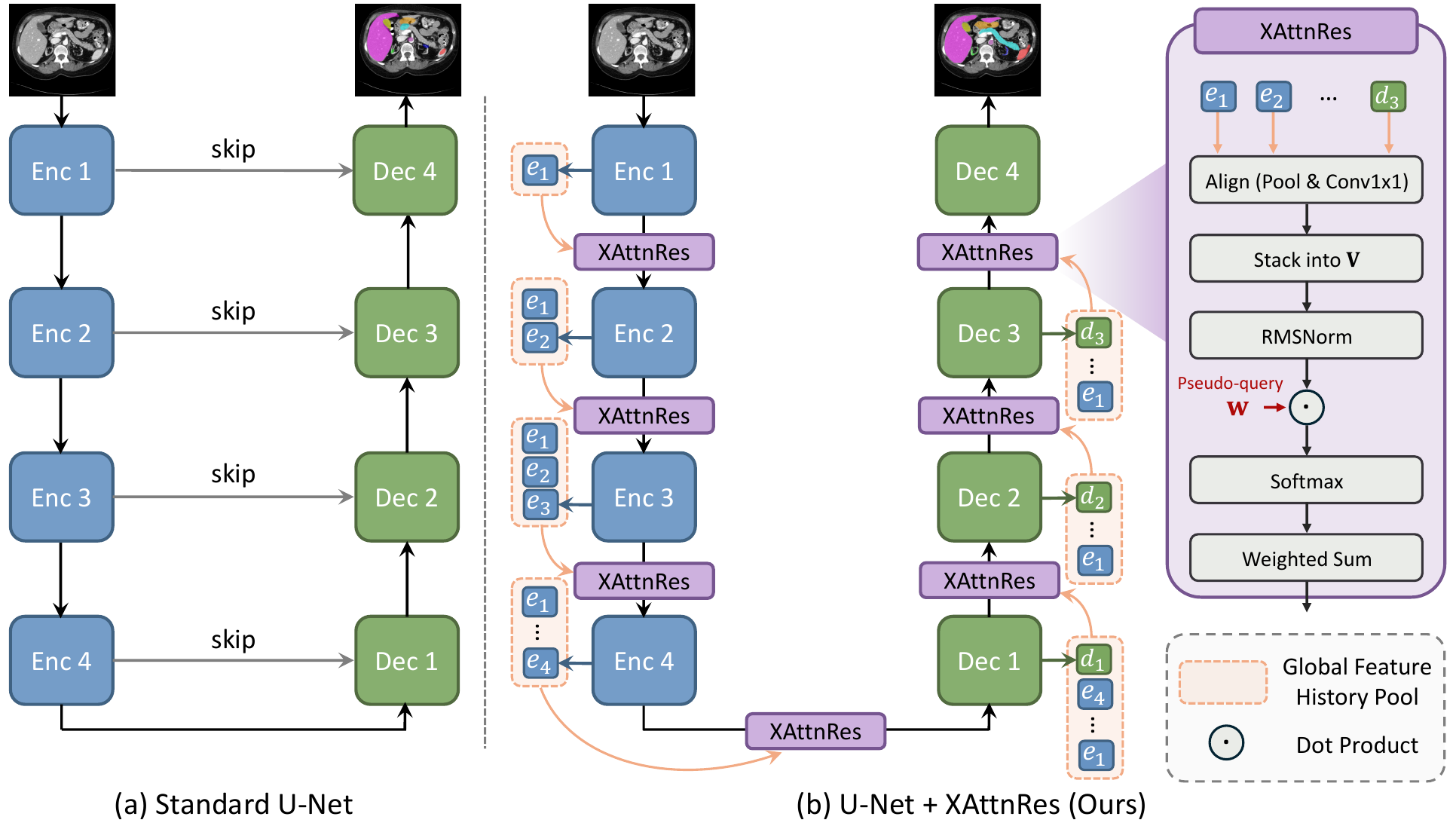}
    \caption{\textbf{Architecture overview.} (a)~Standard U-Net with fixed skip connections between resolution-matched encoder and decoder stages. (b)~U-Net with XAttnRes (replace): skip connections are entirely removed. Each stage reads from a causally growing history pool ($e_1, \ldots, e_S$ for encoder; $e_1, \ldots, e_S, d_1, \ldots$ for decoder) via lightweight pseudo-query attention, and appends its output for subsequent stages. The XAttnRes detail (right) shows how it aligns multi-scale features, computes attention logits via a single learnable vector $\mathbf{w}$, and outputs a weighted sum of the original values.}
    \label{fig:architecture}
\end{figure}

\paragraph{Key insight.}
In a standard U-Net (Figure~\ref{fig:architecture}(a)), information flows between stages through fixed, architecturally predetermined pathways. Attention Residuals~\cite{chen2026attnres} showed that analogous fixed pathways in Transformers (residual connections) can be improved by learned aggregation over all preceding layers. We apply the same principle to encoder-decoder segmentation: XAttnRes adds a learned routing path that aggregates from all preceding stages, providing each stage with data-driven access to the full network history. As we show in our ablation study (Section 4), this learned routing can even fully replace skip connections (Figure~\ref{fig:architecture}(b)) while maintaining performance.

\paragraph{Global feature history pool.}
We maintain an ordered list $\mathcal{H}$ that accumulates the output of every stage during the forward pass. After encoder stage $i$ produces output $e_i$, we append $e_i$ to $\mathcal{H}$. Similarly, after decoder stage $j$ produces $d_j$, we append $d_j$. The history pool thus grows monotonically:
\begin{equation}
    \mathcal{H}_{\text{enc}} = [e_1, e_2, \ldots, e_S], \qquad \mathcal{H}_{\text{dec}} = [e_1, \ldots, e_S,\, d_1, d_2, \ldots],
\end{equation}
where $S$ is the number of encoder stages. Features in $\mathcal{H}$ have heterogeneous shapes: $e_i \in \mathbb{R}^{B \times C_i \times H_i \times W_i}$, with spatial resolution and channel count varying across stages.

\paragraph{Cross-stage alignment.}
Unlike in LLMs where all layer outputs share the same dimensionality, U-Net stages produce features at different resolutions and channel widths. Before applying attention, we align all history features to the target stage's spatial resolution and channel dimension. For a current feature $\mathbf{x} \in \mathbb{R}^{B \times C \times H' \times W'}$, each history feature $h_k$ is aligned via:
\begin{equation}
    \hat{h}_k = \phi_k(h_k) = \text{Conv}_{1\times1}^{(k)}\!\Big(\text{Resize}\big(h_k,\, (H', W')\big)\Big) \in \mathbb{R}^{B \times C \times H' \times W'},
    \label{eq:align}
\end{equation}
where $\text{Resize}(\cdot)$ denotes adaptive max pooling for downsampling or bilinear interpolation for upsampling, and $\text{Conv}_{1\times1}^{(k)}$ projects to $C$ channels. When the source and target channel dimensions match, the convolution reduces to an identity mapping.

\paragraph{Pseudo-query attention aggregation.}
After alignment, we stack all representations into a value tensor:
\begin{equation}
    \mathbf{V} = [\hat{h}_1;\, \ldots;\, \hat{h}_K;\, \mathbf{x}] \in \mathbb{R}^{(K+1) \times B \times C \times H' \times W'}.
\end{equation}
We then normalize $\mathbf{V}$ along the channel dimension to obtain keys, and compute attention logits via dot product with a learnable pseudo-query $\mathbf{w} \in \mathbb{R}^C$ (zero-initialized):
\begin{equation}
    l_n^{(p)} = \mathbf{w}^\top \cdot \text{RMSNorm}(\mathbf{V}_n^{(p)}),
    \label{eq:logits}
\end{equation}
where the superscript $(p)$ indexes a spatial position. The logits are converted to attention weights via softmax over the $K{+}1$ history entries:
\begin{equation}
    \alpha_n^{(p)} = \frac{\exp(l_n^{(p)})}{\sum_{m=1}^{K+1} \exp(l_m^{(p)})}.
    \label{eq:alpha}
\end{equation}
The output is the weighted sum of the original unnormalized values:
\begin{equation}
    \text{XAttnRes}(\mathcal{H},\, \mathbf{x})^{(p)} = \sum_{n=1}^{K+1} \alpha_n^{(p)} \cdot \mathbf{V}_n^{(p)}.
    \label{eq:output}
\end{equation}
The attention operates \emph{independently per spatial position}: organ boundaries may attend strongly to high-resolution encoder features, while homogeneous tissue regions may prefer deeper, more abstract representations.

\paragraph{Integration with encoder-decoder architectures.}
In a standard U-Net, encoder stages are connected by simple downsampling, and decoder stages receive a fixed concatenation of the upsampled previous output and the resolution-matched encoder feature:
\begin{equation}
    \mathbf{x}_i^{\text{enc}} = f_i^{\text{enc}}\!\Big(\text{Down}(\mathbf{x}_{i-1}^{\text{enc}})\Big), \quad
    \mathbf{x}_i^{\text{dec}} = f_i^{\text{dec}}\!\Big([\text{Up}(\mathbf{x}_{i+1}^{\text{dec}})\,;\, \mathbf{x}_i^{\text{enc}}]\Big).
\end{equation}
XAttnRes adds learned aggregation over the history pool as an additional routing path at each stage:
\begin{align}
    \mathbf{x}_i^{\text{enc}} &= f_i^{\text{enc}}\!\Big(\text{XAttnRes}\big(\mathcal{H},\, \text{Down}(\mathbf{x}_{i-1}^{\text{enc}})\big)\Big), \\
    \mathbf{x}_i^{\text{dec}} &= f_i^{\text{dec}}\!\Big(\text{XAttnRes}\big(\mathcal{H},\, \text{Up}(\mathbf{x}_{i+1}^{\text{dec}})\big)\Big).
\end{align}
On the encoder side, each stage aggregates from all preceding encoder representations before processing, rather than receiving only the previous stage's downsampled output. On the decoder side, the decoder retrieves information from any preceding encoder or decoder stage through the pseudo-query mechanism. In our default configuration, XAttnRes operates alongside existing skip connections. As we show in our ablation study (Section~\ref{sec:ablation}), XAttnRes can also fully replace skip connections while maintaining similar performance.

\subsection{Integration as a versatile plug-in}
\label{sec:plugin}

XAttnRes is architecture-agnostic and can be added to any encoder-decoder network as a lightweight plug-in. We demonstrate this with two architectures:

\begin{itemize}
    \item \textbf{U-Net + XAttnRes.} We add XAttnRes to a standard U-Net~\cite{ronneberger2015unet} at each encoder and decoder stage. The encoder and decoder convolution blocks and existing skip connections are unchanged. XAttnRes operates alongside them, providing an additional learned routing path.
    \item \textbf{EMCAD + XAttnRes.} We add XAttnRes to the EMCAD~\cite{rahman2024emcad} decoder alongside its existing LGAG, MSCAM and EUCB components. Since EMCAD employs a pretrained PVT encoder, XAttnRes is only applied on the decoder side, but the history pool accumulates both encoder and decoder stage outputs.
\end{itemize}

\paragraph{Parameter and compute overhead.} Each XAttnRes instance introduces one pseudo-query $\mathbf{w} \in \mathbb{R}^C$, one RMSNorm with $C$ parameters, and $1\times1$ convolution projections for channel alignment (shared when source and target dimensions match). The overhead is modest: as shown in Table~\ref{tab:2d}, UNet + XAttnRes uses 1.28M parameters compared to 1.06M for the baseline (+0.22M overhead), and EMCAD + XAttnRes uses 28.09M compared to 26.76M (+1.33M overhead).

\paragraph{Preservation of high-resolution details.} A natural concern is whether spatial pooling during alignment destroys the fine-grained details that skip connections are designed to preserve. We note that this concern applies only to \emph{cross-resolution} retrieval, i.e., when a high-resolution stage reads from a lower-resolution history entry (requiring upsampling) or vice versa. When a decoder stage retrieves from its resolution-matched encoder stage, no spatial resizing is needed, and the feature is forwarded without any spatial information loss. This is exactly the information pathway that traditional skip connections provide. In practice, the resolution-matched features naturally recover the behavior of skip connections. The cross-resolution pathways provide additional capabilities that standard skip connections lack entirely, which is access to semantic context from other scales. Thus, XAttnRes can in principle recover the behavior of skip connections while also enabling cross-resolution information flow that standard skip connections do not provide.

\section{Experiments}

\subsection{Datasets}

We evaluate on the Synapse multi-organ CT benchmark for multi-class segmentation and three 2D colonoscopy/dermoscopy datasets for binary segmentation:

\begin{itemize}
    \item \textbf{Synapse}~\cite{synapse}: 30 abdominal CT scans (3,779 axial slices) with 8 organ classes. We follow the standard split of 18 training and 12 testing cases \cite{chen2021transunet}.
    \item \textbf{ColonDB}~\cite{tajbakhsh2015colondb}: 380 colonoscopy images for polyp segmentation.
    \item \textbf{ClinicDB}~\cite{bernal2015clinicdb}: 612 colonoscopy images for polyp segmentation.
    \item \textbf{ISIC 2017}~\cite{codella2018isic2017}: 2,750 dermoscopic images for skin lesion segmentation.
\end{itemize}

\noindent For the ColonDB and ClinicDB datasets, we follow the same dataset split as EMCAD~\cite{rahman2024emcad}, partitioning each dataset into training, validation, and test sets with an 8:1:1 ratio. For ISIC 2017, we use the official competition split \cite{codella2018skin}.

\subsection{Implementation details}

We implement all models in PyTorch and conduct experiments on a single NVIDIA RTX A100 GPU (40GB). The training configurations strictly follow those in EMCAD \cite{rahman2024emcad}. All models are trained with the AdamW optimizer using a learning rate and weight decay of $1 \times 10^{-4}$. For the Synapse multi-organ dataset, we train for 300 epochs with batch size 6, resizing images to $224 \times 224$ with random rotation and flipping augmentations, and optimize a combined Cross-Entropy (weight 0.3) and Dice (weight 0.7) loss. For the 2D binary segmentation datasets (ColonDB, ClinicDB, ISIC 2017), we train for 200 epochs with batch size 16, resizing images to $352 \times 352$ with a multi-scale $\{0.75, 1.0, 1.25\}$ training strategy and gradient clipping at 0.5. For EMCAD + XAttnRes, we follow \cite{rahman2024emcad} and use ImageNet-pretrained PVTv2-B2~\cite{wang2022pvtv2} as the encoder. We save the best model based on the Dice score on the validation set.

\begin{table}[!tb]
\caption{Comparison on the Synapse multi-organ CT segmentation dataset. $\uparrow$: higher is better. $\downarrow$: lower is better. Best in \textbf{bold}, second \underline{underlined}.}
\label{tab:synapse}
\centering
\setlength{\tabcolsep}{3.5pt}
\resizebox{\textwidth}{!}{%
\begin{tabular}{l ccc cccccccc}
\toprule
\multirow{2}{*}{Method} & \multicolumn{3}{c}{Average} & \multicolumn{8}{c}{Dice Score (\%)} \\
\cmidrule(lr){2-4} \cmidrule(lr){5-12}
 & DSC$\uparrow$ & HD95$\downarrow$ & mIoU$\uparrow$ & Aorta & GB & KL & KR & Liver & PC & Spleen & SM \\
\midrule
U-Net \cite{ronneberger2015unet} & 76.25 & 26.93 & 65.96 & \underline{89.93} & 54.42 & 78.87 & 72.87 & 92.71 & 59.65 & 85.98 & 75.55 \\
Unet3+ \cite{huang2020unet3p} & 76.93 & 26.77 & 66.02 & 88.87 & 61.93 & 83.06 & 76.52 & 93.83 & 56.21 & 81.88 & 73.17 \\ 
TransUNet \cite{chen2021transunet} & 77.61 & 26.90 & 67.32 & 86.56 & 60.43 & 80.54 & 78.53 & 94.33 & 58.47 & 87.06 & 75.00 \\
Swin-UNet \cite{cao2022swinunet} & 77.58 & 27.32 & 66.88 & 81.76 & 65.95 & 82.32 & 79.22 & 93.73 & 53.81 & 88.04 & 75.79 \\
SSFormer \cite{wang2022ssformer} & 78.01 & 25.72 & 67.23 & 82.78 & 63.74 & 80.72 & 78.11 & 93.53 & 61.53 & 87.07 & 76.61 \\
PolypPVT \cite{dong2023polypvt} & 78.08 & 25.61 & 67.43 & 82.34 & 66.14 & 81.21 & 73.78 & 94.37 & 59.34 & 88.05 & 79.40 \\
MT-UNet \cite{wang2022mtunet} & 78.59 & 26.59 & -- & 87.92 & 64.99 & 81.47 & 77.29 & 93.06 & 59.46 & 87.75 & 76.81 \\
UCTransNet \cite{wang2022uctransnet} & 80.09 & 22.94 & 69.47 & \textbf{90.35} & 61.56 & 82.80 & 77.83 & 94.94 & 62.02 & 90.74 & 80.44 \\
MISSFormer \cite{huang2022missformer} & 81.96 & 18.20 & -- & 86.99 & 68.65 & 85.21 & 82.00 & 94.41 & 65.67 & 91.92 & 80.81 \\
PVT-CASCADE \cite{rahman2023cascade} & 81.06 & 20.23 & 70.88 & 83.01 & \textbf{70.59} & 82.23 & 80.37 & 94.08 & 64.43 & 90.10 & 83.69 \\
TransCASCADE \cite{rahman2023cascade} & 82.68 & 17.34 & 73.48 & 86.63 & 68.48 & 87.66 & \underline{84.56} & 94.43 & 65.33 & 90.79 & 83.52 \\
EMCAD \cite{rahman2024emcad} & \underline{83.63} & \underline{15.68} & \underline{74.65} & 88.14 & \underline{68.87} & \underline{88.08} & 84.10 & \textbf{95.26}& \textbf{68.51} & \underline{92.17} & \underline{83.92} \\
\midrule
\rowcolor{gray!10}
UNet + XAttnRes & 76.40 & 26.71 & 66.17 & 85.20 & 55.81 & 75.92 & 71.73 & 94.96 & 60.50 & 88.71 & 78.34 \\
\rowcolor{gray!10}
\quad $\Delta$ vs U-Net & \textcolor{red}{+0.15} & \textcolor{red}{-0.22} & \textcolor{red}{+0.21} & \textcolor{blue}{-4.73} & \textcolor{red}{+1.39} & \textcolor{blue}{-2.95} & \textcolor{blue}{-1.14} & \textcolor{red}{+2.25} & \textcolor{red}{+0.85} & \textcolor{red}{+2.73} & \textcolor{red}{+2.79} \\
\rowcolor{gray!10}
EMCAD + XAttnRes & \textbf{83.75} & \textbf{15.58} & \textbf{74.85} & 89.80 & 68.00 & \textbf{88.23} & \textbf{84.78} & \underline{95.00} & \underline{66.70} & \textbf{92.74} & \textbf{84.75} \\
\rowcolor{gray!10}
\quad $\Delta$ vs EMCAD & \textcolor{red}{+0.12} & \textcolor{red}{-0.10} & \textcolor{red}{+0.20} & \textcolor{red}{+1.66} & \textcolor{blue}{-0.87} & \textcolor{red}{+0.15} & \textcolor{red}{+0.68} & \textcolor{blue}{-0.26} & \textcolor{blue}{-1.81} & \textcolor{red}{+0.57} & \textcolor{red}{+0.83} \\
\bottomrule
\end{tabular}
}
\end{table}

\subsection{Main results}

\paragraph{Synapse multi-organ segmentation.}
Table~\ref{tab:synapse} presents results on the Synapse multi-organ CT benchmark, comparing XAttnRes with CNN-based methods (U-Net~\cite{ronneberger2015unet}, UNet 3+~\cite{huang2020unet3p}), skip connection improvements (UCTransNet~\cite{wang2022uctransnet}), hybrid CNN-Transformer methods (TransUNet~\cite{chen2021transunet}, MT-UNet~\cite{wang2022mtunet}), pure Transformer methods (Swin-UNet~\cite{cao2022swinunet}, MISSFormer~\cite{huang2022missformer}), and recent decoders (PVT-CASCADE~\cite{rahman2023cascade}, TransCASCADE~\cite{rahman2023cascade}, EMCAD~\cite{rahman2024emcad}).

When our mechanism is applied to a vanilla U-Net (``UNet + XAttnRes''), XAttnRes improves the average Dice from 76.25\% to 76.40\% (+0.15\%) and reduces HD95 from 26.93 to 26.71 mm. At the organ level, five of eight organs show improvement, with the largest gains on Stomach (+2.79\%), Spleen (+2.73\%), and Liver (+2.25\%). This demonstrates that learned cross-stage aggregation provides useful information beyond what standard skip connections already carry.

When integrated into the state-of-the-art EMCAD framework, XAttnRes improves performance from 83.63\% to 83.75\% DSC and from 15.68 to 15.58 mm HD95, confirming that the learned routing generalizes across different architectural designs and benefits even highly optimized decoder pipelines.

\paragraph{2D binary segmentation.}
Table~\ref{tab:2d} evaluates XAttnRes on three 2D binary segmentation datasets spanning two modalities: colonoscopy (ColonDB, ClinicDB) and dermoscopy (ISIC 2017). UNet + XAttnRes achieves consistent improvements across all three datasets, with an average Dice of 86.94\% (+0.18\% over U-Net). EMCAD + XAttnRes reaches 91.28\% average Dice (+0.12\% over EMCAD), obtaining the highest score on all three datasets. These results confirm that the learned routing mechanism generalizes across imaging modalities and is not specific to the multi-class CT setting of Synapse.

\begin{table}[!tb]
\caption{Comparison on 2D binary segmentation datasets. We report \#Params, \#FLOPs, and Dice score (\%) on ColonDB, ClinicDB (polyp), and ISIC 2017 (skin lesion). Best in \textbf{bold}, second \underline{underlined}.}
\label{tab:2d}
\centering
\setlength{\tabcolsep}{4pt}
\resizebox{\linewidth}{!}{\begin{tabular}{l cc ccc c}
\toprule
Method & \#Params & \#FLOPs & ColonDB & ClinicDB & ISIC 2017 & Average \\
\midrule
U-Net \cite{ronneberger2015unet} & 1.06M & 2.50G & 86.01 & 92.13 & 82.13 & 86.76 \\
UNet++ \cite{zhou2018unetpp} & 9.16M & 34.65G & 87.88 & 92.17 & 82.98 & 87.68 \\
Attn U-Net \cite{oktay2018attention} & 34.88M & 66.64G & 86.46 & 92.20 & 83.66 & 87.44 \\
DeepLabv3+ \cite{chen2018deeplabv3p} & 39.76M & 14.92G & 91.92 & 93.24 & 83.84 & 89.67 \\
PraNet \cite{fan2020pranet} & 32.55M & 6.93G & 89.16 & 91.71 & 83.03 & 87.97 \\
Unet3+ \cite{huang2020unet3p} & 26.97M & 199.74G & 88.38 & 92.60 & 83.34 & 88.11 \\
CaraNet \cite{lou2022caranet} & 46.64M & 11.48G & 91.19 & 94.08 & 85.02 & 90.10 \\
TransUNet \cite{chen2021transunet} & 105.32M & 38.52G & 91.63 & 93.90 & 85.00 & 90.18 \\
SwinUNet \cite{cao2022swinunet} & 27.17M & 6.20G & 89.27 & 92.42 & 83.97 & 88.55 \\
UCTransUnet \cite{wang2022uctransnet} & 66.49M & 43.06G & 91.68 & 93.24 & 84.13 & 89.68 \\
PVT-CASCADE \cite{rahman2023cascade} & 34.12M & 7.62G & 91.60 &  94.53 & 85.50 & 90.54 \\
EMCAD \cite{rahman2024emcad} & 26.76M & 5.60G & \underline{92.31} & \underline{95.21} & \underline{85.95} & \underline{91.16} \\
\midrule
\rowcolor{gray!10}
UNet + XAttnRes & 1.28M & 2.87G & 86.22 & 92.35 & 82.25 & 86.94 \\
\rowcolor{gray!10}
\quad {$\Delta$ vs U-Net} & +0.22M & +0.37G & \textcolor{red}{+0.21} & \textcolor{red}{+0.22} & \textcolor{red}{+0.12} & \textcolor{red}{+0.18} \\
\rowcolor{gray!10}
EMCAD + XAttnRes & 28.09M & 5.76G & \textbf{92.40} & \textbf{95.32} & \textbf{86.13} & \textbf{91.28} \\
\rowcolor{gray!10}
\quad {$\Delta$ vs EMCAD} & +1.33M & +0.16G & \textcolor{red}{+0.09} & \textcolor{red}{+0.11} & \textcolor{red}{+0.18} & \textcolor{red}{+0.12} \\
\bottomrule
\end{tabular}}
\end{table}

\begin{table}[!tb]
\caption{Ablation: skip connection vs.\ XAttnRes across two backbones and two benchmarks. ``Replace'' removes all skip connections and uses only XAttnRes. ``Both'' retains skip connections and adds XAttnRes on top.}
\label{tab:ablation_skip}
\centering
\setlength{\tabcolsep}{4pt}
\begin{tabular}{lcc ccc}
\toprule
\multirow{2}{*}{Model} & \multirow{2}{*}{Skip} & \multirow{2}{*}{XAttnRes} & \multicolumn{2}{c}{Synapse} & ColonDB \\
\cmidrule(lr){4-5} \cmidrule(lr){6-6}
 & & & DSC$\uparrow$ & HD95$\downarrow$ & DSC$\uparrow$ \\
\midrule
U-Net (baseline) & \checkmark & -- & 76.25 & 26.93 & 86.01 \\
U-Net (no skip) & -- & -- & 72.70 & 31.44 & 83.28 \\
U-Net + XAttnRes (replace) & -- & \checkmark & 76.31 & 26.95 & 86.07 \\
U-Net + XAttnRes (both) & \checkmark & \checkmark & 76.40 & 26.71 & 86.22 \\
\midrule
EMCAD (baseline) & \checkmark & -- & 83.63 & 15.68 & 92.31 \\
EMCAD (no skip) & -- & -- & 75.47 & 28.97 & 89.81 \\
EMCAD + XAttnRes (replace) & -- & \checkmark & 83.45 & 15.73  & 92.28 \\
EMCAD + XAttnRes (both) & \checkmark & \checkmark & 83.75 & 15.58 & 92.40 \\
\bottomrule
\end{tabular}
\end{table}

\subsection{Ablation studies}
\label{sec:ablation}

We conduct ablation experiments on the Synapse and ColonDB benchmarks to analyze the design choices of XAttnRes.

\paragraph{Skip connection vs.\ XAttnRes.}
Table~\ref{tab:ablation_skip} compares four information routing configurations on two backbones. Adding XAttnRes alongside skip connections (``both'') yields the best results across all settings (U-Net: 76.40\%; EMCAD: 83.75\% on Synapse), confirming that learned routing provides useful information beyond what skip connections carry. Removing skip connections without any replacement (``no skip'') causes a large performance drop on both backbones (U-Net: 76.25$\to$72.70; EMCAD: 83.63$\to$75.47), confirming that inter-stage information flow is critical. An interesting finding is that XAttnRes alone (``replace''), with all skip connections removed, is able to recover this drop and approach the skip connection baseline on Synapse. This suggests that the information traditionally carried by predetermined skip connections can also be recovered through learned aggregation over the feature history pool, making XAttnRes a promising alternative to fixed skip topologies.

\begin{table}[!tb]
\centering
\begin{minipage}[t]{0.48\textwidth}
\centering
\caption{Application position on Synapse.}
\label{tab:ablation_pos}
\resizebox{\linewidth}{!}{%
\begin{tabular}{l cc}
\toprule
Position & DSC (\%)$\uparrow$ & HD95$\downarrow$ \\
\midrule
None & 72.70 & 31.44 \\
Encoder only & 74.09 & 29.50 \\
Decoder only & 75.62 & 28.02 \\
Full (Enc. + Dec.) & 76.40 & 26.71 \\
\bottomrule
\end{tabular}
}
\end{minipage}
\hfill
\begin{minipage}[t]{0.48\textwidth}
\centering
\caption{Pseudo-query initialization on Synapse.}
\label{tab:ablation_init}
\resizebox{\linewidth}{!}{%
\begin{tabular}{l cc}
\toprule
Initialization & DSC (\%)$\uparrow$ & HD95$\downarrow$ \\
\midrule
Zero-init & 76.40 & 26.71 \\
Random & 75.74 & 27.82 \\
Kaiming uniform & 74.32 & 29.20 \\
Xavier uniform & 75.55 & 28.07 \\
\bottomrule
\end{tabular}
}
\end{minipage}
\end{table}

\begin{figure}[!tb]
\centering
\includegraphics[width=0.95\linewidth]{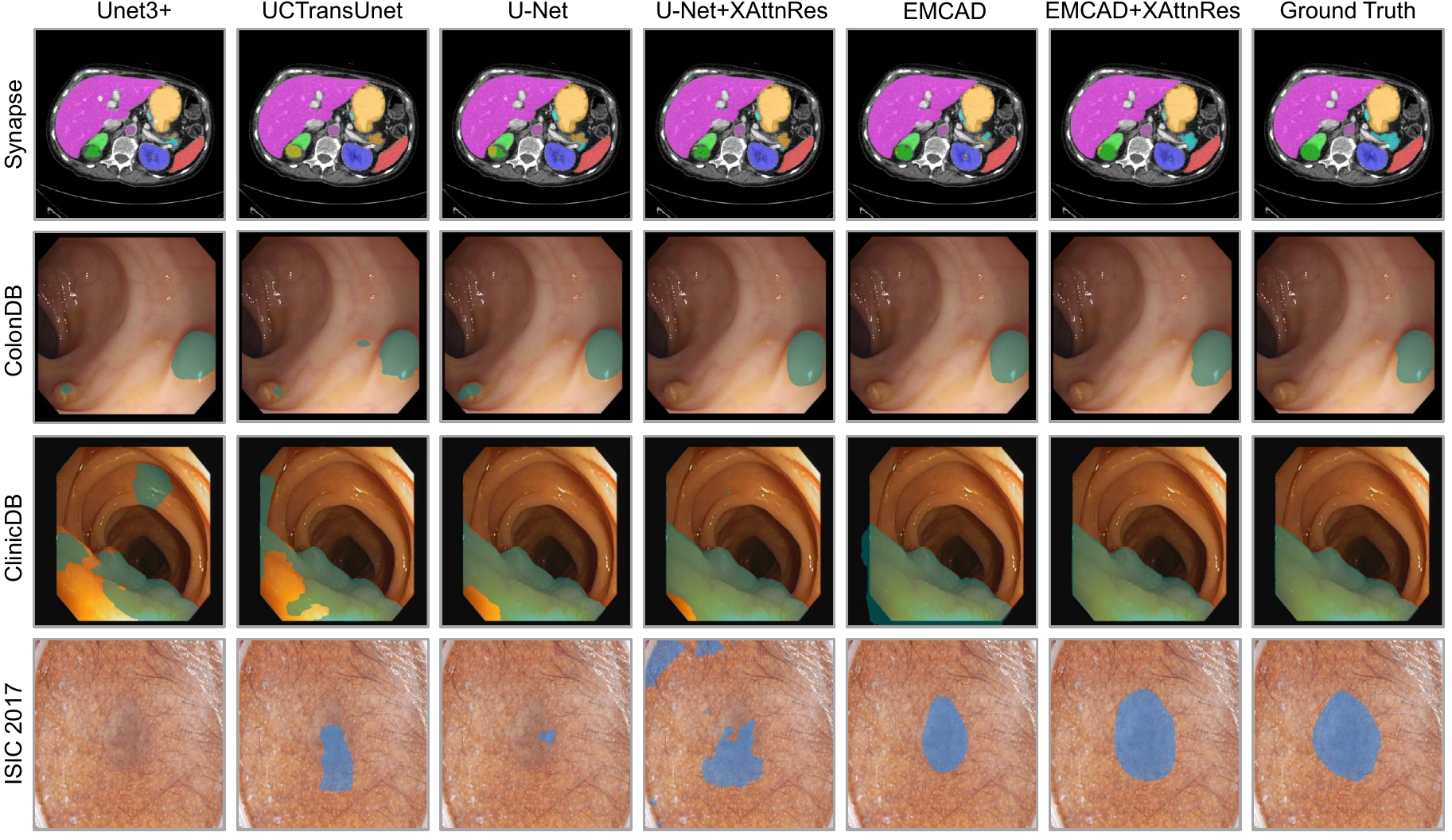}
\caption{Qualitative comparison across four datasets. Each row shows one dataset (Synapse, ColonDB, ClinicDB, ISIC 2017). Columns from left to right: UNet 3+, UCTransNet, U-Net, U-Net + XAttnRes (ours), EMCAD, EMCAD + XAttnRes (ours), and ground truth.}
\label{fig:qualitative}
\end{figure}

\paragraph{Application position.}
Table~\ref{tab:ablation_pos} studies where to apply XAttnRes within the U-Net backbone (without skip connections). Applying XAttnRes to the decoder alone recovers most of the skip connection performance (75.62\% vs.\ 76.25\% baseline), which is expected since decoder-side aggregation serves the same functional role as traditional skip connections: it provides the decoder with access to encoder representations. Encoder-only application yields a smaller gain (74.09\%), as it enables cross-stage feature reuse within the encoder but does not address the encoder-to-decoder information bottleneck. The full configuration (both encoder and decoder, 76.40\%) is additive, confirming that learned aggregation benefits both paths.

\paragraph{Pseudo-query initialization.}
Table~\ref{tab:ablation_init} examines the initialization strategy for the pseudo-query vectors. Zero initialization achieves the best result (76.40\%), outperforming random normal (75.74\%, $-$0.66\%), Xavier uniform (75.55\%, $-$0.85\%), and Kaiming uniform (74.32\%, $-$2.08\%). Zero initialization causes the mechanism to start as uniform averaging over all history entries and gradually specialize during training. This conservative starting point avoids the risk of early training instability from random attention patterns, and aligns with the finding in the original AttnRes work~\cite{chen2026attnres} for LLMs. The sensitivity to initialization is notable: Kaiming uniform, a standard choice for ReLU networks, performs worst here, likely because it produces large initial logits that create sharp, premature routing decisions.

\subsection{Qualitative visualization}

Figure~\ref{fig:qualitative} presents qualitative segmentation results across all four datasets. On the Synapse dataset (top row), U-Net + XAttnRes produces organ masks with improved completeness compared to the vanilla U-Net. On ColonDB and ClinicDB (middle rows), XAttnRes variants maintain tight polyp boundaries. On ISIC 2017 (bottom row), both U-Net + XAttnRes and EMCAD + XAttnRes capture the irregular contours of skin lesions. These visualizations are consistent with the quantitative improvements observed in Tables~\ref{tab:synapse} and~\ref{tab:2d}.

\section{Conclusion}

We have presented Cross-Stage Attention Residuals (XAttnRes), the first adaptation of Attention Residuals from large language models to medical image segmentation. By maintaining a global feature history pool and employing lightweight pseudo-query attention, XAttnRes enables each stage to selectively aggregate information from \emph{all} preceding encoder and decoder representations, with the routing learned purely from data. When added alongside existing skip connections, XAttnRes consistently improves performance across four datasets and three imaging modalities. We further find that XAttnRes alone, without any skip connections, can match the performance of traditional skip connection baselines, suggesting that learned aggregation is a promising direction for feature routing in encoder-decoder segmentation. We hope this work encourages further exploration of learned routing mechanisms in medical image analysis.


%
%
\bibliographystyle{splncs04}
\bibliography{main}
\end{document}